\def\year{2022}\relax
\documentclass[letterpaper]{article} 
\usepackage{aaai22}  
\usepackage{times}  
\usepackage{helvet}  
\usepackage{courier}  
\usepackage[hyphens]{url}  
\usepackage{graphicx} 
\usepackage{natbib}  
\usepackage{caption} 
\DeclareCaptionStyle{ruled}{labelfont=normalfont,labelsep=colon,strut=off} 
\usepackage{algorithm}
\usepackage{algorithmic}

\usepackage{newfloat}
\usepackage{listings}
\floatstyle{ruled}
\newfloat{listing}{tb}{lst}{}
\floatname{listing}{Listing}
\usepackage{multirow}
\usepackage{array}
\usepackage{fancyhdr}
\usepackage{booktabs}
\usepackage{amsmath}
\usepackage{amsfonts}

\title{Confidence Calibration for Intent Detection \\ via Hyperspherical Space and Rebalanced Accuracy-Uncertainty Loss}

\author{
  Yantao Gong$^{1,2,3}$, 
  Cao Liu$^3$, 
  Fan Yang$^3$, 
  Xunliang Cai$^3$, \\
  Guanglu Wan$^3$, 
  Jiansong Chen$^3$, 
  Weipeng Zhang$^3$, 
  Houfeng Wang$^{2, \dagger}$
}

\affiliations{
  $^1$ School of Software and Microelectronics, Peking University \\
  $^2$ MOE Key Lab of Computational Linguistics, Peking University 
  $^3$ Meituan \\
  \{gongyt, wanghf\}@pku.edu.cn
  \{liucao, yangfan79, caixunliang, wanguanglu, chenjiansong, zhangweipeng02\}@meituan.com
}

\usepackage{bibentry}

\begin{document}

\maketitle

\renewcommand{\thefootnote}{\fnsymbol{footnote}}
\footnotetext[2]{Corresponding author.}

\begin{abstract}
    Data-driven methods have achieved notable performance on intent detection, which is a task to comprehend user queries. Nonetheless, they are controversial for over-confident predictions. 
    In some scenarios, users do not only care about the accuracy but also the confidence of model.  
    Unfortunately, mainstream neural networks are poorly calibrated, with a large gap between accuracy and confidence. 
    To handle this problem defined as confidence calibration, we propose a model using the hyperspherical space and rebalanced accuracy-uncertainty loss. 
    Specifically, we project the label vector onto hyperspherical space uniformly to generate a dense label representation matrix, which mitigates over-confident predictions due to overfitting sparce one-hot label matrix. 
    Besides, we rebalance samples of different accuracy and uncertainty to better guide model training. 
    Experiments on the open datasets verify that our model outperforms the existing calibration methods and achieves a significant improvement on the calibration metric. 
\end{abstract}

\section{Introduction}

Intent detection is a crucial portion in comprehending user queries, which generally predicts intent tags by semantic classification \cite{brenes2009survey, Qin2020AGIFAA}. Therefore, it is widely used in many NLP applications, such as search, task-based dialogue, and other fields \cite{Zhang2016AJM,larson2019evaluation,casanueva2020efficient}.

\begin{figure}[t]
  \centering
    \includegraphics[width=0.9\columnwidth]{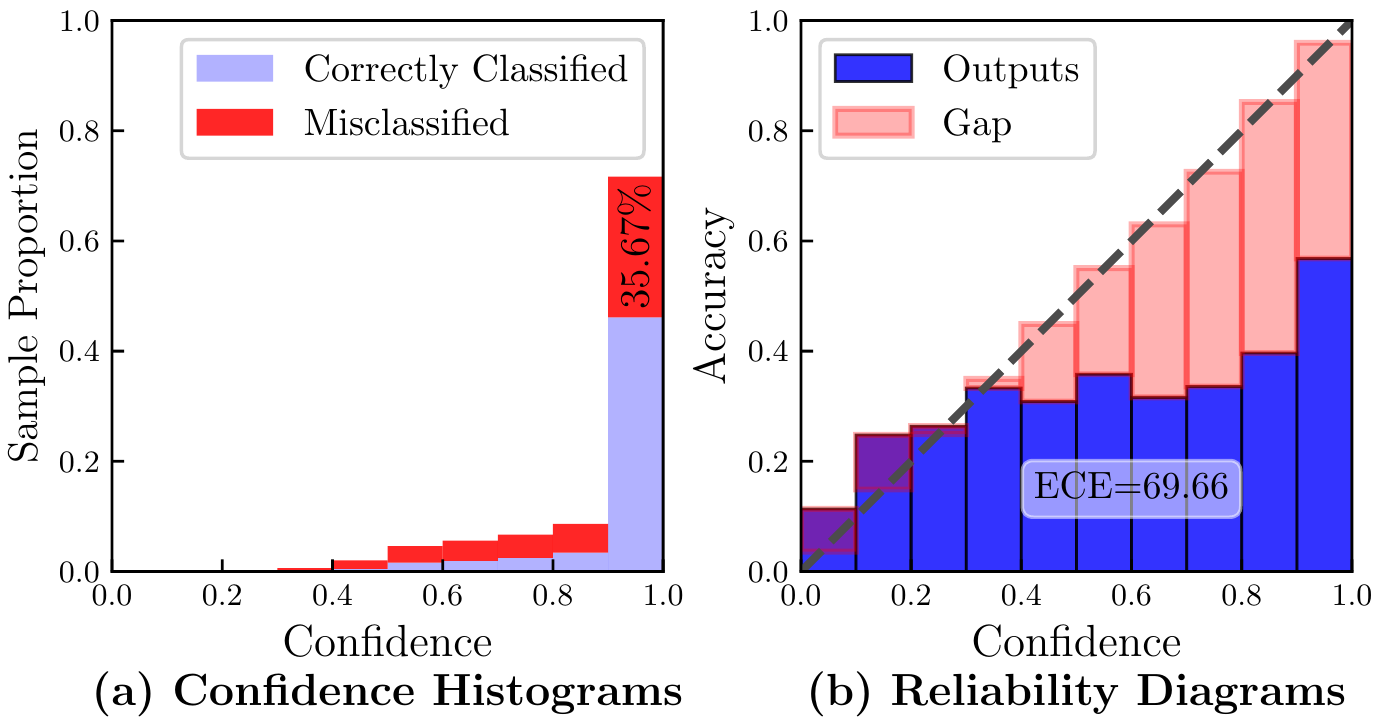}
    \caption{
    The confidence histograms and reliability diagrams of TNEWS dataset. We employ the fine-tuned BERT model to perform statistical analysis. As the figure demonstrates, when the confidence is between 0.9 and 1, misclassified samples constitute about 35.67\%. ``Gap" represents the difference between confidence and accuracy. Model is worse calibrated if the ``Gap" is larger. Fine-tuned BERT model without calibration tends to make over-confident predictions and possesses a high \textit{expected calibration error} (ECE). 
    }
    \label{fig:introduct}
\end{figure}

In recent years, data-driven methods develop rapidly and become a primary trend of intent detection. However, they are highly criticized for over-confident predictions \cite{NiculescuMizil2005PredictingGP, Nguyen2015DeepNN, Pereyra2017RegularizingNN, Li2020RegularizationVS}. 
As shown in Figure \ref{fig:introduct}(a), 
there is a serious phenomenon that the prediction confidence (i.e. probability associated with the predicted label) of samples is very high even if the samples are misclassified. 
For example, when the confidence is in [0.9-1], the proportion of misclassified samples reaches 35.67\%. 
Besides, the average confidence (90.39\%) is evidently over the accuracy (56.17\%). 


One of the effective solutions to deal with the aforementioned problem is confidence calibration. 
A perfectly calibrated model is supposed to output average confidence equal to the accuracy \cite{kong-etal-2020-calibrated, Kppers2020MultivariateCC}. 
Unfortunately, due to over-parameterization and overfitting of the conventional methods, mainstream neural networks are poorly calibrated \cite{krishnan2020improving, wang2020inference, Schwaiger2021FromBT, Enomoto2021LearningTC}. 
As demonstrated in Figure \ref{fig:introduct}, ``Gap" means the discrepancy between the average confidence and accuracy. The larger the ``Gap" as, the worse the model is calibrated. Model without calibration, indicated in Figure \ref{fig:introduct}(b), 
easily faces under-estimation problem when confidence is less than 0.2 and over-estimation problem when confidence is more than 0.4. 
Therefore, it owns a higher \textit{expected calibration error} (ECE, calibration metric, more details in Section \ref{sec:ece}) than perfectly calibrated model. 

\begin{figure}
  \centering
    \includegraphics[width=0.9\columnwidth]{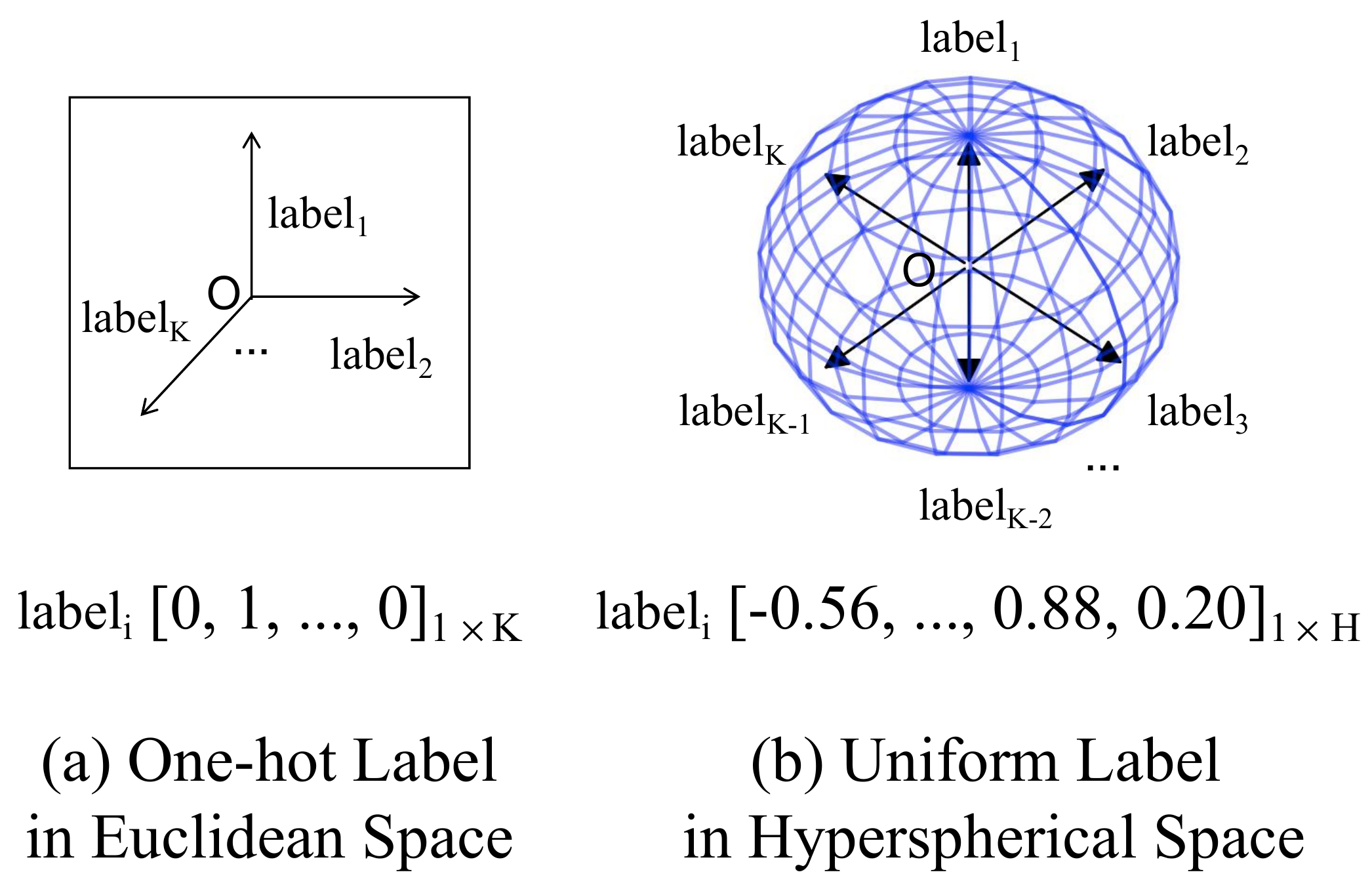}
    \caption{
      Representation of label vectors in euclidean space and hyperspherical space. 
      One-hot label vectors are in the form of a sparse matrix and only use the positive portion. 
      Additionally, one-hot label vectors require at least as many dimensions as the size of label set like $K$ ($K\ll H$ in most cases). 
      In contrast, label vectors in hyperspherical space are shaped into a dense matrix and employ the entire portion without dimension limitation. 
    }
    \label{fig:label_vector}
\end{figure}

To handle the confidence calibration problem, researchers have proposed numerous works \cite{Nguyen2015PosteriorCA, szegedy2016rethinking, Mller2019WhenDL}. 
One primary calibration approach acts on the post-processing stage. \citeauthor{guo2017calibration} \shortcite{guo2017calibration} provide temperature scaling, which learns a single parameter from the development dataset to rescale all the logit before transmitting to softmax. 
Another way to calibrate the model is by designing a particular loss function to minimize the discrepancy between accuracy and confidence. 
\citeauthor{krishnan2020improving} \shortcite{krishnan2020improving} lately propose the accuracy versus uncertainty calibration loss (AVUC loss), which leverages the relationship between accuracy and uncertainty as an anchor for calibration, and it obtains a significant improvement. 

Nevertheless, the aforementioned methods have some important issues. 
1) As demonstrated in Figure \ref{fig:label_vector}, one of the problems lies in that the above methods project the labels in the form of a one-hot matrix in Euclidean space, which is sparse and merely uses the positive portion of the output space.  
During the training process, such a sparse matrix is easy to bring about the network to make over-confident predictions, as proved by \citeauthor{Szegedy2016RethinkingTI} \shortcite{Szegedy2016RethinkingTI} and \citeauthor{Mller2019WhenDL} \shortcite{Mller2019WhenDL}. 
2) Another issue is that although \citeauthor{krishnan2020improving} \shortcite{krishnan2020improving} divide samples into several groups according to their accuracy and uncertainty, it treats accurate and inaccurate samples equally. 
In fact, there exists a large number of misclassified samples with high confidence (low uncertainty), displayed in Figure \ref{fig:introduct}(a), which suggests that the model is misleading by the wrong signal during training. 

In order to deal with the above issues, we propose a model employing the \textbf{H}yperspherical \textbf{S}pace and \textbf{R}ebalanced \textbf{A}ccuracy-\textbf{U}ncertainty loss (HS-RAU) to process confidence calibration for the intent detection task. 
Specifically, 1) We project the label vector onto the hyperspherical space uniformly, as vividly shown in Figure \ref{fig:label_vector}. 
Hyperspherical space uses a dense matrix to represent labels and employs the entire portion of the output space rather than one-hot labels.  
In this way, we mitigate the overfitting problem of model to the sparse one-hot matrix. 
2) We propose a rebalanced accuracy-uncertainty loss to capitalize on the properties of distinct samples. 
Through RAU loss, we optimize the accurate samples with high uncertainty and the inaccurate samples with low uncertainty respectively, which contributes to better guide model training. 

To validate the effectiveness of our model, we conduct abundant experiments on the three open datasets. 
Empirical results demonstrate that our model achieves evident improvements compared with the SOTA. Specifically, F1 increases on all the datasets with the calibration metric (ECE) drops down 10.50\% on average. On the TNEWS dataset, the ECE achieves an obvious amelioration of 29.67\% and the F1 obtains 1.21\% promotion. Furthermore, our model acquires better performance among the existing methods on noisy data and low-frequency labels.

To sum up, our contributions are as follows: 

(1) We uniformly project the label vectors onto the hyperspherical space to obtain a denser representation matrix, which mitigates the model to overfit the sparce one-hot label matrix and generate over-confident predictions. 

(2) We rebalance the accuracy and uncertainty of samples  
and optimize the accurate samples with low uncertainty and inaccurate samples with high uncertainty separately by RAU loss to provide better guidance in the training process. 

(3) The experimental results demonstrate that our model gains an advantage over the SOTA, not only in the F1 but also in the confidence calibration metric. 
Moreover, we obtain noteworthy performance on noisy data and low-frequency labels. 

\begin{figure*}[t]
  \centering
    \includegraphics[width=0.8\textwidth]{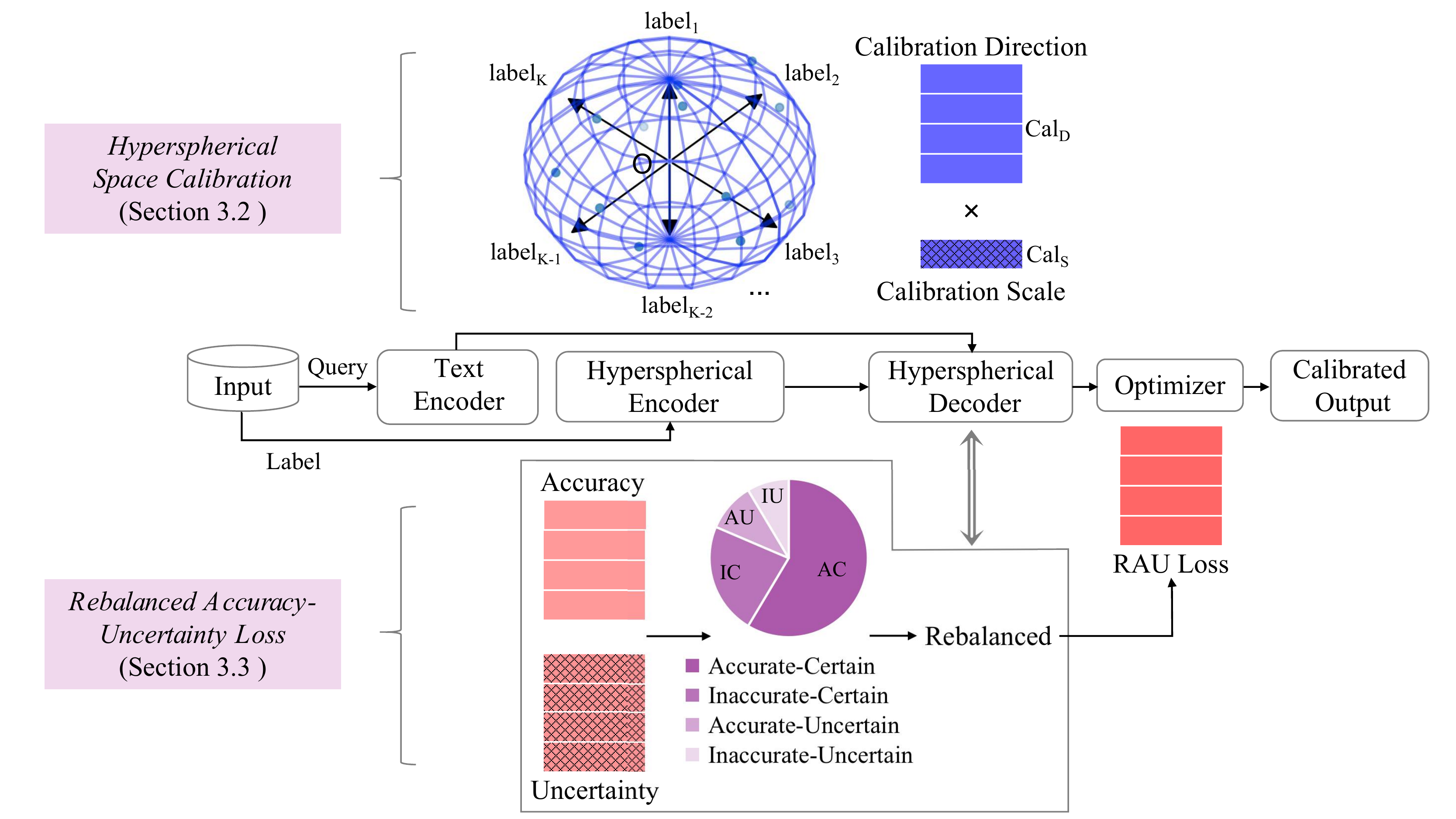}
    \caption{The illustration of confidence calibration via Hyperspherical Space and Rebalanced Accuracy-Uncertainty loss (HS-RAU) framework. 
    After getting the encoded vector of input query by the text encoder, we project the label vector onto the hyperspherical space uniformly, and encode the input label by the hyperspherical encoder to obtain a dense label matrix. Then, we compute the calibration direction matrix as well as the calibration scale. 
    Next, we partition the samples into four sets according to their accuracy and uncertainty, rebalance the samples' accuracy and uncertainty by the RAU loss. Through the above process, we acquire the output with calibration.  
    }
    \label{fig:model}
\end{figure*}

\section{Related Work}
\textbf{Intent Detection.}
Intent is the sematic purpose of a query, which is generated by users \cite{Xu2013ConvolutionalNN, Wang2018AttentionBasedCN}. As a matter of fact, the essence of intent detection is text classification \cite{brenes2009survey, Mehri2020ExampleDrivenIP, Chatterjee2020IntentMF}. 
After training on the dataset with ground-truth labels, the model attempts to predict the intent of query within the existing intent set. 
There have been plenty of researches on conventional neural network methods in the last few decades \cite{Xu2013ConvolutionalNN, Liu2016AttentionBasedRN,Zhang2019JointSF,Haihong2019ANB, Wang2020MaskedfieldPF, Gerz2021MultilingualAC}. 
During recent years, with the rapid development of computing power, pre-trained models such as BERT \cite{devlin2018bert} are employed for intent detection frequently \cite{Castellucci2019MultilingualID, He2019UsingCN, Zhang2019AJL, Athiwaratkun2020AugmentedNL, Gong2021DensityBasedDC}. 

\noindent\textbf{Confidence Calibration.}
Confidence calibration has a long history of research in statistical machine learning \cite{brier1950verification, Griffin1992TheWO, Gneiting2007StrictlyPS}. In the past several years, one major calibration methods fall into the post-processing stage \cite{Platt1999ProbabilisticOF, Zadrozny2001ObtainingCP, Kumar2019VerifiedUC, Zhang2020MixnMatchEA, rahimi2020intra}. 
For example, \citeauthor{guo2017calibration} \shortcite{guo2017calibration} propose the temperature scaling. The trained model learns a single calibration scale from the development set. 
Another main calibration approaches try to optimize a function that represents the difference of average confidence and accuracy \cite{Kumar2018TrainableCM, Kull2019BeyondTS, Mukhoti2020CalibratingDN, Gupta2020CalibrationON}. 
For instance, \citeauthor{krishnan2020improving} \shortcite{krishnan2020improving} devise a loss function to combine accuracy and uncertainty. \citeauthor{Jung2020PosteriorCT} \shortcite{Jung2020PosteriorCT} come up with a method to minimize the distribution between predicted probability and empirical probability.

\section{Method}
\subsection{Model Overview}
As shown in Figure \ref{fig:model}, we employ the \textit{hyperspherical space and rebalanced accuracy-uncertainty loss} to process confidence calibration. 
First, we use a text encoder such as BERT to acquire the encoded vector of the input query. 
Next, through the hyperspherical encoder, we obtain the dense encoded matrix of the input labels, which alleviates the over-confident predictions caused by the overfitting of sparse one-hot label matrix. 
After that, we utilize the hyperspherical decoder to calculate the calibration direction matrix and calibration scale. 
Furthermore, we separate the samples according to their accuracy and uncertainty, and design the rebalanced accuracy-uncertainty loss to optimize accurate and inaccurate samples respectively. In the end, we obtain the output with calibration.

\subsection{Hyperspherical Space Calibration}
In this submodule, 
we introduce how to separate the hyperspherical space homogeneously and project label vectors onto the hyperspherical space to obtain a dense label matrix. 
\subsubsection*{Text Encoder}
For $N$ queries \{$Q_1, ..., Q_i, ..., Q_N$\}, 
the corresponding labels are \{$T_1, ..., T_i, ..., T_N$\}, where $T_i \in C$. 
$C = \{1, ..., K\}$ indicates the set of $K$ label tags. 
We exploit the text encoder like BERT to extract the encoded vector $E_i$ ($H$ dimension) such as [CLS] of each input query $Q_i$. The encoded vector matrix $E$ of all the queries is calibrated in the hyperspherical decoder. 

\subsubsection*{Hyperspherical Encoder}
Before the learning process, we separate the H-dimensional output space $\mathbb{S}^{H}$ into $K$ subspaces uniformly, 
which has the same size as the label set $C$. 
Then, we define the vector of the hyperspherical label as \{$h_1, ..., h_i, ..., h_K$\}, corresponding to the $K$ subspaces. 
In addition, the norm of each vector satisfies $\left \| h_i \right \| = 1 $. 
The dimension of hyperspherical label vector is $H$, which equals the dimension of encoded vector.  
The hyperspherical encoder encodes each input label to a dense hyperspherical label vector, 
which is utilized in the hyperspherical decoder for calibration. 

Here comes the detail of uniformly projecting the label vectors onto hyperspherical space. 
For each label vector $h_i$ in the hyperspherical space, it has $K-1$ cosine distances between all the $K-1$ label vectors except itself, and the max cosine distance among them is $D_i$, defined as below: 
\begin{equation}
  D_i = \max_{} (d_{ij})
\end{equation}
where $i, j \in C \ $and$  \ i\ne j$. $d_{ij}$ is the cosine distance between label vector $h_i$ and $h_j$. 
As our goal is to make the label vector uniformly distributed in the hyperspherical space, therefore, it is equivalent to the optimization problem that minimizes the sum of the maximum cosine distance $D_i$ of each label vector, as the following modality: 
\begin{equation}
  \mathcal{L}_h = \min\frac{1}{K} \sum_{i=1}^{K} D_i
\end{equation}

Furthermore, due to all the label vectors are unit vectors, the above formula can be converted to matrix multiplication, which speeds up the calculation, by the following equations: 
\begin{equation}
  \begin{split}
  \mathcal{L}_h = \min\frac{1}{K} \sum_{i=1}^{K} \max_{}  (Z_i), \\
  Z = X \cdot X^{T}-2I ~~~~~~\\
  \end{split}
\end{equation}
where $X = \begin{bmatrix}h_1, ..., h_i, ..., h_K \end{bmatrix}$ is the matrix of hyperspherical label vector. $I$ is the identity matrix. 
$Z_i$ is the $i^{th}$ row of $Z$. In order to avoid self-selection, $Z$ subtracts identity matrix $I$ twice. 

\subsubsection*{Hyperspherical Decoder}

After acquiring the encoded query vector and the dense encoded hyperspherical label vector through hyperspherical encoder, we utilize the hyperspherical decoder to get the calibration direction and the calibration scale. 

We perform the dot product of the encoded vector with each hyperspherical label vector to get the calibration direction matrix $Cali_D$, formulated as below: 
\begin{equation}
  Cali_D = E\cdot X^{T}
\end{equation}
where $E\in \mathbb{R} ^{N\times H} $ denotes the encoded vector matrix of all the queries. $X^{T}\in \mathbb{R} ^{H\times K} $ is the transpose matrix of dense hyperspherical label vector. 
Then, we calculate the norm of label vector matrix as the calibration scale $Cali_S$, which is the scale parameter during the overall process, by using the following equation: 
\begin{equation}
Cali_S=\begin{Vmatrix} X
 \end{Vmatrix}
\end{equation}

Finally, we compute the calibrated new logit $L$ as below: 
\begin{equation}
  L =  Cali_S  \times Cali_D
\end{equation}
where the calibration scale $Cali_S$ is an unidimensional variable and the calibration direction $Cali_D\in \mathbb{R} ^{N\times K} $.

\subsection{Rebalanced Accuracy-Uncertainty Loss}
\label{sec:rebalanced}
In this submodule, we design the rebalanced accuracy-uncertainty loss to optimize accurate and inaccurate samples separately. 
Whether a sample is considered as accurate depends on whether the predicted label of the sample ${T_i}'$ equals to the exact sample's label $T_i$, so we define the confidence (probability of predicted label) of a single sample as $a_i$ in the following: 
\begin{equation}
  a_i=\begin{cases}
    \max_{} (p_i),& \text{ if } {T_i}'=T_i \\
    1-\max_{} (p_i),& \text{ otherwise. }
  \end{cases}
\end{equation}
where $p_i$ is the predicted probability after transmitting to softmax. 
Therefore, when the predictions are accurate the $a_i$ is close to 1, while it is close to 0 when inaccurate. 
As there is no ground truth evaluation of the uncertainty, we utilize the calculation method described in \citeauthor{krishnan2020improving} \shortcite{krishnan2020improving} to get the uncertainty $u_i$ as follows:  
\begin{equation}
  u_i = -p_i\log_{}{p_i} 
\end{equation}

Then, we set the uncertainty threshold as $u_{\theta } \in \left [ 0,1 \right ]$, which is a heuristic setting obtained through the average uncertainty of training samples from initial epochs. 
A sample is defined as certain when the uncertainty of it is lower than $u_{\theta }$. Otherwise, it's defined as uncertain. 
Then, we divide the training samples into four sets \{$AC, AU, IC, IU$\} separately, where $AC$ means Accurate-Certain, $AU$ means Accurate-Uncertain, $IC$ means Inaccurate-Certain, and $IU$ means Inaccurate-Uncertain. 

Based on the assumption mentioned in \citeauthor{krishnan2020improving} \shortcite{krishnan2020improving}, a well-calibrated model provides a low uncertainty for accurate predictions while it provides a high uncertainty for inaccurate predictions.
Therefore, the model with calibration is supposed to produce a higher $AVU \in \left [ 0,1 \right ]$ measure. $AVU$ is computed by summing the number of \{$AC, IU$\} two sets, and then divide the total number of \{$AC, AU, IC, IU$\} four sets.  

To make the $AVU$ function differentiable for neural network parameters, we devise the calculation methods like:
\begin{equation}
  \begin{split}
  n_{AC} = {\textstyle \sum_{i\in \left \{ {T_i}' = T_i \ and\ u_i\le u_\theta  \right \}}^{}}  a_i\odot (1- \tan (u_i)), \\
  n_{AU} = {\textstyle \sum_{i\in \left \{ {T_i}' = T_i \ and\ u_i>u_\theta  \right \}}^{}}  a_i\odot \tan (u_i),  ~~~~~~~~~\\
  n_{IC} = {\textstyle \sum_{i\in \left \{ {T_i}' \ne  T_i \ and\ u_i\le u_\theta  \right \}}^{}} a_i\odot (1- \tan (u_i)), \\
  n_{IU} = {\textstyle \sum_{i\in \left \{ {T_i}' \ne  T_i \ and\ u_i> u_\theta  \right \}}^{}} a_i\odot \tan (u_i) ~~~~~~~~~~~\\
  \end{split} 
\end{equation}
where $\odot$ is hadamard project. 
In addition, we step further on and rebalance the accuracy-uncertainty, which prompts the model to respectively optimize accurate samples with low uncertainty and inaccurate samples with high uncertainty during training. 
To be specific, we define the RAU loss as: 
\begin{equation}
  \mathcal{L}_{RAU}=\log_{}{( 1 + \frac{n_{AU}}{n_{AC}+n_{AU}} + \frac{n_{IC}}{n_{IC}+n_{IU}})}  
\end{equation}

When $n_{AU}$ and $n_{IC}$ are optimized close to zero, the RAU loss is close to zero, which means the model is certain about the predictions of accurate samples, while there are no over-confident predictions of the inaccurate samples. 

\begin{table*}\footnotesize
  \setlength{\tabcolsep}{4.5pt}
  \centering
  \renewcommand\arraystretch{1.15}
  \begin{tabular}{ccccccccc}
  \toprule[1.2pt]
  \multirow{2.5}{*}{Model}   & \multicolumn{2}{c}{TNEWS} & \multicolumn{2}{c}{HWU64} & \multicolumn{2}{c}{BANKING77} & \multicolumn{2}{c}{Average} \\  \cmidrule(lr){2-3} \cmidrule(lr){4-5} \cmidrule(lr){6-7} \cmidrule(lr){8-9}
                      & F1     & ECE     & F1     & ECE     & F1       & ECE       & F1      & ECE      \\  \midrule[1.2pt]
                      BERT \cite{devlin2018bert}               & 54.81            & 69.66           & 91.85            & 17.18           & 93.61              & 11.98             & 80.09             & 32.94            \\
                      TS \cite{guo2017calibration}                & 54.81            & 49.88           & 91.85            & 15.86           & 93.61              & 11.87             & 80.09             & 25.87            \\
                      LS \cite{Mller2019WhenDL}                & 55.29            & 53.99           & 92.06            & 16.51           & 93.86              & 11.40             & 80.40             & 27.30            \\
                      PosCal \cite{Jung2020PosteriorCT}            & 54.98            & 68.05           & 92.03            & 16.14           & 93.66              & 11.93             & 80.30             & 32.05            \\
                      AVUC \cite{krishnan2020improving}              & 55.41            & 67.98           & 92.02            & \textbf{15.71}  & 93.83              & 11.79             & 80.42             & 31.83            \\ \midrule
                      HS-RAU (Ours) & \textbf{56.02}   & \textbf{39.99}  & \textbf{92.52}   & 16.12  & \textbf{93.89}     & \textbf{11.21}    & \textbf{80.81}    & \textbf{22.44}   \\ \midrule[1.2pt]
  \end{tabular}
  \caption{Overall comparison with different calibration methods on three open datasets.}
  \label{tab:main_res}
  \end{table*}

\section{Experiments}
\subsection{Experimental Setup}
\subsubsection{Experimental Datasets}
We mainly experiment on three open datasets described below. The download links are displayed in Appendix A. 

\noindent\textit{TNEWS}, a Chinese dataset proposed by \citeauthor{xu2020clue} \shortcite{xu2020clue}, has identical essence with intent detection. It includes 53360 samples in 15 categories. The provided test set are without gold labels. So we regard validation set as test set and randomly 
divide 5000 samples from training set for validation. 

\noindent\textit{HWU64}, proposed by \citeauthor{Liu2019BenchmarkingNL} \shortcite{Liu2019BenchmarkingNL} to reflects human-home robot interaction, which owns 15726 samples spanning 64 intents. We use one fold train-test split with 9960 training samples and 1076 testing samples. 

\noindent\textit{BANKING77}, proposed by \citeauthor{casanueva2020efficient} \shortcite{casanueva2020efficient}, which has 13083 samples, 9002 for training and 3080 for testing. This dataset consists of 77 intents in a single domain of online banking inquiry. 

\subsubsection{Comparison Methods}
We compare with the methods as listed below: 

\noindent\textit{BERT} \cite{devlin2018bert}: represents the pre-trained base BERT model. 

\noindent\textit{Temperature Scaling (TS)} \cite{guo2017calibration}: is the classical post-processing method learning a single parameter from the dev dataset to rescale the logit after the model is trained. 

\noindent\textit{Label Smoothing (LS)} \cite{Mller2019WhenDL}: smoothes some part of the one-hot label' probability to a weighted mixture probability of the none ground-truth labels, which is set to compare our hyperspherical labels.  

\noindent\textit{Posterior Calibrated (PosCal)} \cite{Jung2020PosteriorCT}: minimizes the difference between the predicted and empirical posterior probabilities, which is a competitive recent research.  

\noindent\textit{Accuracy Versus Uncertainty Calibration (AVUC)} \cite{krishnan2020improving}: proposes an optimization method that utilizes the relevance of accuracy and uncertainty as an anchor for calibration. 

\subsubsection{Implementation Details}
All experiments are taken on BERT (with or without confidence calibration) unless otherwise specified. 
We employ Adam \cite{Kingma2015AdamAM} as the optimizer and search learning rate in \{4e-5, 5e-5\} with the training epochs in \{19, 23\} and about 40s per epoch. 
To make full use of the GPU memory, we set the batch size to 256. The type of GPU is Tesla V100. 
Besides, the KL loss between predicted probability and empirical probability is added optionally in PosCal, AVUC, and our model. 
More implementation details are shown in Appendix A. 

\subsubsection{Confidence Calibration Metric}
\label{sec:ece}
We follow the previous researches and utilize 
the \textit{expected calibration error} (ECE) \cite{Naeini2015ObtainingWC}, 
which is a common evaluation metric to calculate the calibration error in confidence calibration. ECE separates the predictions of samples into $M$ bins according to the predicted probability called confidence. Then, accumulating the weighted differences between accuracy and confidence in each bin:   
\begin{equation}
  ECE =\frac{1}{K}\sum_{i=1}^{K}\sum_{j=1}^{M }\frac{\left |B_{ij} \right |}{N}  \left | Acc_{ij}-Con_{ij}  \right |   
\end{equation}
where $\left |B_{ij} \right |$ is the size of bin $j$ in label $i$, $N$ is the number of total prediction samples, $Acc_{ij}$ is the empirical probability and $Con_{ij}$ is the average predicted probability for label $i$ in bin $j$ respectively.   

\begin{figure*}[h]
  \centering
    \includegraphics[width=0.8\textwidth]{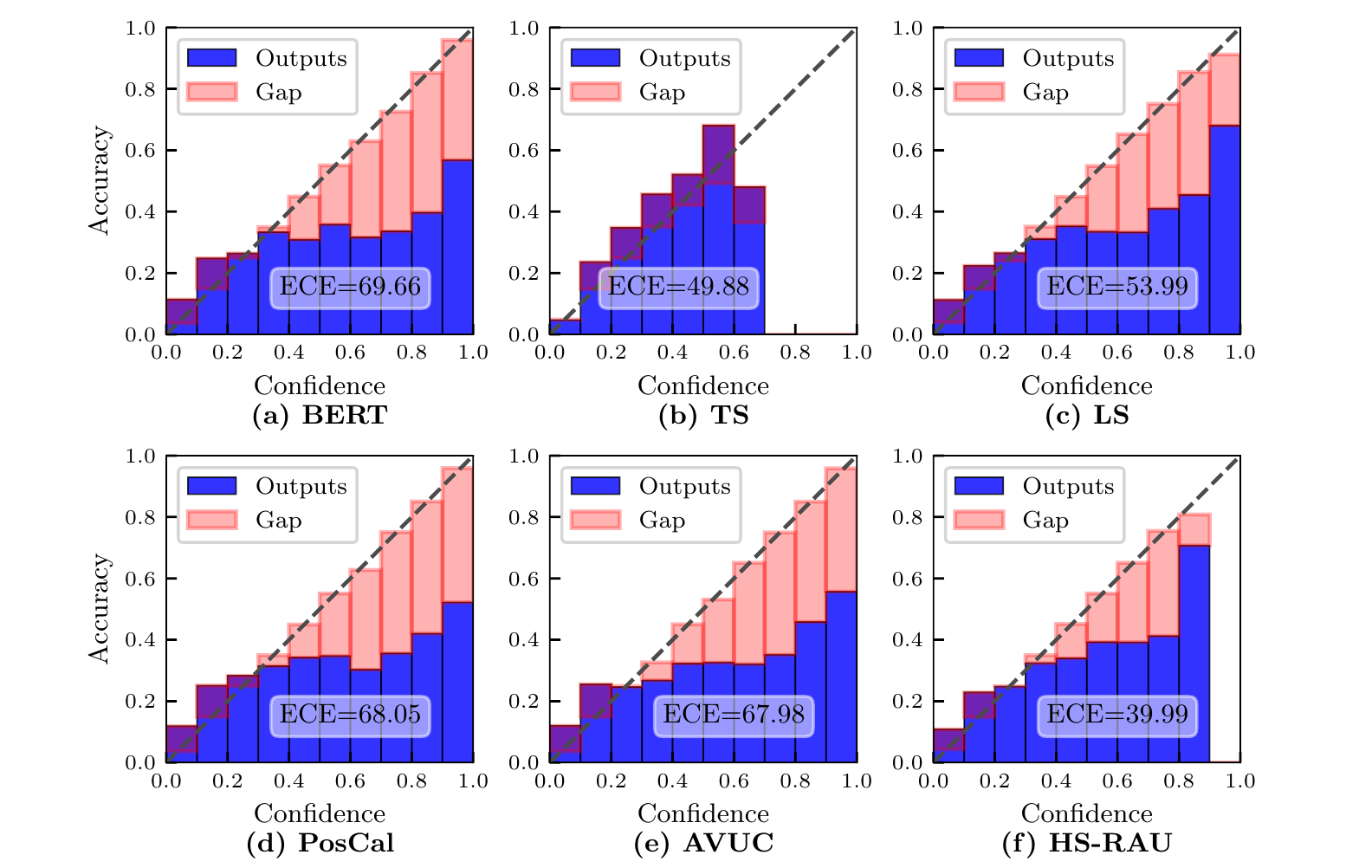}
    \caption{Reliability diagrams for TNEWS dataset, before calibration (a) and after calibration (b)-(f).
    }
    \label{fig:reliability}
\end{figure*}

  \subsection{Comparison with State-of-the-arts}

  \textbf{Comparison Settings.} 
  We reproduce the baselines, and the results are almost equal to the published metrics. Based on that, 
  we conduct extensive experiments on the above three datasets to validate the effectiveness of our model. F1 and ECE are considered as the main evaluation metrics. 
  
  \textbf{Comparison Results.} 
  As Table \ref{tab:main_res} illustrated: 
  
  (1) Regardless of which dataset we choose, our model achieves the best F1 performance among all the calibration methods. Specially, we obtain a significant reduction in ECE, which drops down 10.50\% on average compared with the baseline model. This proves the effectiveness of our model to project label vector onto hyperspherical space uniformly and utilize the rebalanced accuracy-uncertainty loss in confidence calibration.  
  
  (2) All the calibration methods have limited amelioration in the dataset possessing better performance. It makes sense for the reason that the model is well studied on these datasets. Hence the main distribution of its confidence is as high as the accuracy, like 90+\%, which results in more credible predictions. 
  So we majorly analyze the TNEWS dataset in the subsequent experiments. 
  
  (3) In the case of TNEWS dataset, the F1 gains 1.21\% over BERT while the ECE decreases remarkably. 
  Furthermore, more information on the rest datasets can be inquired in Appendix B.


  \subsection{Observing Miscalibration}
  
  \textbf{Comparison Settings.}
  We use the reliability diagrams for observation, which is a visual representation of model calibration \cite{NiculescuMizil2005PredictingGP}.
  The average confidence within each bin is defined as ``Outputs", while the absolute difference between confidence and accuracy in each bin is defined as ``Gap". 
  The ECE is proportional to ``Gap" in some degree, described in Sec. \ref{sec:ece}. 
  
  \textbf{Comparison Results.}
  As Figure \ref{fig:reliability} depicts, 
  although distinct calibration methods still have a miscalibration phenomenon on TNEWS dataset,  
  BERT with calibration can acquire a lower ECE. 
  Especially, our model decreases ECE prominently and 
  less inclined to make overconfident predictions for samples with higher accuracy
  compared with the AVUC, which manifests the validity of our RAU loss that rebalances accuracy-uncertainty and optimizes accurate as well as inaccurate samples respectively. 
  
  \subsection{Ablation Study}
    \begin{table}[]\footnotesize
      \setlength{\tabcolsep}{4.5pt}
      \centering
      \renewcommand\arraystretch{1.15}
      \begin{tabular}{p{1.60cm}<{\centering}p{0.80cm}<{\centering}p{0.80cm}<{\centering}p{0.80cm}<{\centering}p{0.80cm}<{\centering}p{0.80cm}<{\centering}}
        \toprule[0.9pt]
      \multicolumn{1}{c}{Model} & ACC            & P              & R              & F1             & ECE            \\  \midrule[0.9pt]
      HS-RAU                & \textbf{56.39} & \textbf{56.31} & \textbf{55.82} & \textbf{56.02} & \textbf{39.99} \\ \midrule
      w/o HS                      & 56.32          & 55.82          & 55.26          & 55.52          & 67.97          \\
      w/o RAU                     & 55.90          & 56.05          & 54.75          & 55.28          & 42.53          \\
      w/o Both                  & 56.37          & 56.13          & 54.81          & 55.31          & 68.50          \\  
      \scalebox{0.8}{RAU$\Rightarrow$AVUC}      & 56.21          & 55.72          & 55.45          & 55.55          & 41.82           \\  \midrule[0.9pt]
      \end{tabular}
      \caption{Ablation study on TNEWS dataset by removing the main components, where ``w/o" means without, ``HS" represents hyperspherical space calibration, and ``RAU" indicates rebalanced accuracy-uncertainty loss.}
      \label{tab:ablation}
      \end{table}
      
  \begin{table*}\footnotesize
    \setlength{\tabcolsep}{4.5pt}
    \renewcommand\arraystretch{1.15}
    \centering
    \begin{tabular}{ccccccccc}
      \toprule[1.2pt]
      \multirow{2.5}{*}{Model}                                                           & \multicolumn{2}{c}{5\% Noisy Labels} & \multicolumn{2}{c}{10\% Noisy Labels} & \multicolumn{2}{c}{30\% Noisy Labels} & \multicolumn{2}{c}{50\% Noisy Labels} \\     \cmidrule(lr){2-3} \cmidrule(lr){4-5} \cmidrule(lr){6-7} \cmidrule(lr){8-9} 
      & F1                & ECE              & F1                & ECE              & F1                & ECE              & F1                & ECE              \\  \midrule[1.2pt]
    BERT                                                                                      & 53.46            & 72.19            & 51.44             & 72.38            & 43.76             & 77.98            & 32.22             & 83.35            \\
    LS                                                                                       & 53.95            & 46.46            & 52.59             & 43.81            & 44.83             & 46.92            & 32.91             & 65.94            \\
    PosCal                                                                                   & 54.11            & 52.77            & 52.14             & 61.92            & 45.40             & 56.83            & 33.93             & 71.37            \\
    AVUC                                                                                     & 54.04            & 64.47            & 51.94             & 61.78            & 45.74             & 56.44            & 33.74             & 71.14            \\  \midrule
    HS-RAU                                                                             & \textbf{54.43}   & \textbf{34.37}   & \textbf{53.66}    & \textbf{35.48}   & \textbf{47.03}    & \textbf{32.25}   & \textbf{35.20}    & \textbf{41.06}   \\  \midrule[1.2pt] 
    \end{tabular}
    \caption{Performance on TNEWS dataset with noise.}
    \label{tab:noise}
    \end{table*}
  
  \textbf{Comparison Settings.}
  To validate the effectiveness of our model components, we gradually get rid of some components, including hyperspherical space and rebalanced accuracy-uncertainty loss. 
  In practice, if hyperspherical space is not employed in the model, we use the typical one-hot vector in Euclidean space to represent the labels. 
  
  \textbf{Comparison Results.}
  As described in Table \ref{tab:ablation}: 
  
  (1) Taking out any components of our model results in performance reduction, which certifies the validity of all components. 
  
  (2) Specifically, replacing hyperspherical space leads to conspicuous performance degradation. It shows that projecting labels onto hyperspherical space to obtain a dense label matrix and setting the calibration scale throughout the process can effectively draw down the ECE. 
  
  (3) Substituting RAU for AVUC causes a minor decline in F1, while ECE gets 1.83\% worse. This proves that RAU, which respectively optimizing accurate and inaccurate samples, is beneficial to improve performance. 
  See more results on other datasets in Appendix C. 
  
  \section{Discussion}
  
  \subsection{Effectiveness on Noisy Data}
  \textbf{Comparison Settings.}
  Model performance under noisy data is an important indicator to measure robustness, as it's a frequent phenomenon for data to have noise. 
  We randomly mislabel 5\%, 10\%, 30\%, and 50\% part of samples' labels on TNEWS dataset to simulate the noisy environment. 
  
  \textbf{Comparison Results.}
  The experimental results of 
  Table \ref{tab:noise} support the statements as below: 
  
  (1) The experimental results indicate that our model still obtains significant improvement irrespective of how much noise label is in the dataset. 
  
  (2) Take TNEWS dataset with 30\% error labels as an example, F1 increases 3.27\%. In the meantime, ECE decreases by 45.73\%. 
  During the experiment, we find that although temperature scaling obtains a comparable ECE, the scale parameter turns extremely large, and the output probability is only distributed into one bin. The reason may be that temperature scaling is not suitable for the situation where training set and other sets are labeled differently, as it learns the scale parameter from the development set after the model is trained on the training set. 
  
  (3) Above results verify the effectiveness of our model on the noisy data, which projects the label vector uniformly onto the hyperspherical space and 
  makes better use of the dense label representations. 
  Hyperspherical label vector is not utterly orthogonal like the one-hot label, so mislabeled samples are more likely to be calibrated. 
  More information on other metrics can be found in Appendix D. 
  
  \subsection{Effectiveness on Low-Frequency Labels}
  
  \begin{table}\footnotesize
    \setlength{\tabcolsep}{4.5pt}
    \centering
    \renewcommand\arraystretch{1.15}
    \begin{tabular}
      {p{1.40cm}<{\centering}p{0.80cm}<{\centering}p{0.80cm}<{\centering}p{0.80cm}<{\centering}p{0.80cm}<{\centering}p{0.80cm}<{\centering}}
      \toprule[0.9pt]
    \multirow{2}{*}{Model} & L-F$_{1}$       & L-F$_{2}$       & L-F$_{3}$       & \multicolumn{2}{c}{Average}    \\ \cmidrule(lr){2-6}
                           & F1             & F1             & F1             & F1             & ECE           \\  \midrule[0.9pt]
    BERT                   & 40.54          & 44.50          & 57.54          & 47.53          & 1.61          \\
    TS                    & 40.54          & 44.50          & 57.54          & 47.53          & 3.05          \\
    LS                    & 40.45          & 46.56          & 59.50          & 48.84          & 1.14          \\
    PosCal                & 35.44          & 46.15          & 60.00          & 47.20          & 1.50          \\
    AVUC                  & 43.37          & \textbf{48.31} & \textbf{60.88} & 50.85          & 1.46          \\  \midrule
    HS-RAU                   & \textbf{54.12} & 47.22          & 59.06          & \textbf{53.47} & \textbf{1.08}  \\ \midrule[0.9pt]
    \end{tabular}
    \caption{Performance of the low-frequency labels on TNEWS dataset, where L-F$_{1}$ means the Lowest Frequency label and ``Average" means the average performance of the three lowest frequency labels.}
    \label{tab:LF}
    \end{table}
  
  \textbf{Comparison Settings.}
  Class-imbalanced datasets commonly face the long tail problem. 
  To examine the performance of minority labels, we experiment on the three lowest frequency labels on TNEWS dataset, which contains fifteen classes in total. 
  The sum of fifteen labels' ECE equals the gross ECE. 
  Besides, L-F$_{1}$ means the Lowest Frequency label, L-F$_{2}$ means the second lowest, and so on. 
  ``Average" means the average performance of the three lowest frequency labels.

  \textbf{Comparison Results.}
  As demonstrated in Table \ref{tab:LF}, our model reaches the highest average F1 of the low-frequency labels, with a 5.94\% absolute increase. 
  Apart from that, it works better on the average ECE too. According to the consequences, we can infer that it's no picnic to learn sparse label features (like one-hot) with low-frequency samples. In contrast, the label features of our model are dense, as we separate the label vector into the hyperspherical space horizontally and use more portion of the output space. 
  More details on the comparison results are shown in Appendix E. 
  
  \subsection{Effectiveness on Different Encoders}
  
  \textbf{Comparison Settings.}
  Different encoders have distinct output spaces. To assess the performance of different encoders, we also horizontally compare with other encoders such as Albert-tiny \cite{Lan2020ALBERTAL}, XLNet \cite{Yang2019XLNetGA}, and BERT-large \cite{devlin2018bert} on TNEWS dataset.
  
  \textbf{Comparison Results.}
  The consequences of different encoders on TNEWS dataset are shown in Table \ref{tab:encoder}. 
  Though different encoders behave diversely from each other, they all acquire a comparable enhancement with the help of our confidence calibration strategy. Results indicate that projecting the output space onto hyperspherical space by our strategy possesses a certain universality, which is not limited to the BERT model. 
  
  \begin{table}\footnotesize
    \setlength{\tabcolsep}{4.5pt}
    \centering
    \renewcommand\arraystretch{1.15}
    \begin{tabular}
      {p{1.60cm}<{\centering}p{0.80cm}<{\centering}p{0.80cm}<{\centering}p{0.80cm}<{\centering}p{0.80cm}<{\centering}p{0.80cm}<{\centering}}
      \toprule[0.9pt]
      Model                                    & ACC            & P              & R              & F1             & ECE            \\  \midrule[0.9pt]
    Albert-tiny                           & 52.00          & 48.67          & 48.41          & 48.49          & 28.00          \\
    + HS-RAU                           & \textbf{53.03} & \textbf{50.24} & \textbf{49.18} & \textbf{49.58} & \textbf{23.91} \\  \midrule
    XLNet                                 & 56.84          & \textbf{56.10} & 55.72          & 55.81          & 40.56          \\
    + HS-RAU                           & \textbf{57.06} & \textbf{56.10} & \textbf{56.14} & \textbf{56.04} & \textbf{35.79} \\  \midrule
    BERT-large                            & 57.84          & \textbf{57.59}          & 56.45          & 56.91          & 69.26          \\ 
    + HS-RAU                           & \textbf{58.09} & 57.33 & \textbf{56.65} & \textbf{56.93} & \textbf{39.02} \\  \midrule[0.9pt]
    \end{tabular}
    \caption{Performance of distinct encoders on TNEWS.}
    \label{tab:encoder}
    \end{table}

  \section{Conclusion}
  In this work, we propose a confidence calibration 
  model for intent detection via \textbf{H}yperspherical \textbf{S}pace and \textbf{R}ebalanced \textbf{A}ccuracy-\textbf{U}ncertainty loss (HS-RAU). 
  With the help of projecting label vectors onto hyperspherical space uniformly, we make better use of the dense label representation matrix to mitigate the over-confident predictions as well as the whole portion of output space. 
  Through the rebalanced accuracy-uncertainty loss, we better guide the model to respectively optimize the accurate and inaccurate samples. 
  Experimental results indicate that our model obtains a decent rise over SOTA. Especially, we achieve a significant improvement in the confidence calibration metric (ECE) among the calibration methods.

  \section*{Acknowledgments}
  The work is supported by National Natural Science Foundation of China (Grant No.62036001) and PKU-Baidu Fund (No. 2020BD021).

\bibliography{aaai22.bib}




\clearpage
\appendix

\begin{table*}[h]
  \centering
  \scalebox{0.7}{\setlength{\tabcolsep}{1mm}{
  \begin{tabular}{cccccccccccccccccc}
    \toprule[0.9pt]
  \multirow{2.5}{*}{Model}   & \multicolumn{5}{c}{TNEWS}                                                          & \multicolumn{5}{c}{HWU64}                                                          & \multicolumn{5}{c}{BANKING77}                                                      & \multicolumn{2}{c}{Average} \\ \cmidrule(lr){2-6} \cmidrule(lr){7-11} \cmidrule(lr){12-16} \cmidrule(lr){17-18}
                      & ACC            & P              & R              & F1    & ECE   & ACC            & P              & R              & F1    & ECE   & ACC            & P              & R              & F1    & ECE   & F1       & ECE     \\ \midrule[0.9pt]
  BERT           & 56.17          & 55.64          & 54.31          & 54.81          & 69.66          & 92.01          & 91.64          & 92.66          & 91.85          & 17.18          & 93.60          & 93.85          & 93.60          & 93.61          & 11.98          & 80.09             & 32.94            \\
  TS                 & 56.17          & 55.64          & 54.31          & 54.81          & 49.88          & 92.01          & 91.64          & 92.66          & 91.85          & 15.86          & 93.60          & 93.85          & 93.60          & 93.61          & 11.87          & 80.09             & 25.87            \\
  LS                 & \textbf{56.54} & 55.42          & 55.23          & 55.29          & 53.99          & 92.01          & 92.03          & 92.77          & 92.06          & 16.51          & 93.86          & 94.08          & 93.86          & 93.86          & 11.40          & 80.40             & 27.30            \\
  PosCal             & \textbf{56.54} & 55.46          & 54.66          & 54.98          & 68.05          & 92.10          & 91.96          & 92.73          & 92.03          & 16.14          & 93.67          & 93.90          & 93.67          & 93.66          & 11.93          & 80.30             & 32.05            \\
  AVUC               & 56.21          & 55.81          & 55.13          & 55.41          & 67.98          & 92.19          & 91.76          & 92.90          & 92.02          & \textbf{15.71} & 93.83          & 94.06          & 93.83          & 93.83          & 11.79          & 80.42             & 31.83            \\ \midrule
  HS-RAU (Ours) & 56.39          & \textbf{56.31}          & \textbf{55.82} & \textbf{56.02} & \textbf{39.99} & \textbf{92.57} & \textbf{92.41} & \textbf{93.26} & \textbf{92.52} & 16.12 & \textbf{93.90} & \textbf{94.13} & \textbf{93.90} & \textbf{93.89} & \textbf{11.21} & \textbf{80.81}    & \textbf{22.44}   \\ \bottomrule[0.9pt]
  \end{tabular}}}
  \caption{Overall comparison with different calibration methods on three open datasets.}
  \label{tab:main_append}  
\end{table*}

\begin{table*}[ht]
  \centering
  \scalebox{0.8}{\setlength{\tabcolsep}{1mm}{
  \begin{tabular}{cccccccccccccccc}
    \toprule[0.9pt]
  \multirow{2.5}{*}{Model} & \multicolumn{5}{c}{TNEWS}                                                          & \multicolumn{5}{c}{HWU64}                                                          & \multicolumn{5}{c}{BANKING77}                                                      \\  \cmidrule(lr){2-6} \cmidrule(lr){7-11} \cmidrule(lr){12-16}
                         & ACC            & P              & R              & F1             & ECE            & ACC            & P              & R              & F1             & ECE            & ACC            & P              & R              & F1             & ECE            \\  \midrule[0.9pt]
  HS-RAU             & \textbf{56.39} & \textbf{56.31} & \textbf{55.82} & \textbf{56.02} & \textbf{39.99} & \textbf{92.57} & \textbf{92.41} & \textbf{93.26} & \textbf{92.52} & \textbf{16.12} & \textbf{93.90} & \textbf{94.13} & \textbf{93.90} & \textbf{93.89} & \textbf{11.21} \\  \midrule
  w/o HS                   & 56.32          & 55.82          & 55.26          & 55.52          & 67.97          & 92.01          & 91.66          & 92.66          & 91.78          & 16.30          & 93.83          & 94.03          & 93.83          & 93.82          & 11.75          \\
  w/o RAU                  & 55.90          & 56.05          & 54.75          & 55.28          & 42.53          & 91.82          & 91.67          & 92.60          & 91.83          & 16.57          & 93.70          & 93.93          & 93.70          & 93.69          & 11.93          \\
  w/o Both             & 56.37          & 56.13          & 54.81          & 55.31          & 68.50          & 92.10          & 91.59          & 92.74          & 91.87          & 16.26          & 93.64          & 93.88          & 93.64          & 93.65          & 11.91         \\ \bottomrule[0.9pt]
  \end{tabular}}}
  \caption{Ablation study on three open datasets by removing the main components, where "w/o" means without, "HS" represents hyperspherical space calibration, and "RAU" indicates rebalanced accuracy-uncertainty loss.}
  \label{tab:ablation_append}  
\end{table*}

\section{More Implementation Details}
We employ Adam as the optimizer and search learning rate in \{4e-5, 5e-5\}. The max sequence length is set in \{33, 64\} according to the queries' average length of dataset. Besides,
we set the training epochs in \{19, 23\}. The weight of RAU loss is set 3 and the weight of optionally added KL loss is set 1, tuing from the experiments. The best performance on TNEWS dataset with the learning rate of 4e-5 and 19 epochs. 
HWU64 dataset with the learning rate of 4e-5 and 19 epochs. Besides, it owns a weight decay of 0.05. 
BANKING77 dataset with the learning rate of 5e-5 and 23 epochs. It also owns a weight decay of 0.05. 

Here comes the hyperparameter for other calibration methods. For label smoothing, we set smoothing parameter as 0.1. For PosCal, we search the number of updating empirical probability per epoch in \{1, 3, 5\}. For AVUC, we set the weight of AVUC loss as 3. 

The download link of the three open datasets are as follows: 

(1) TNEWS: \url{https://github.com/CLUEbenchmark/CLUE}

(2) HWU64: \url{https://github.com/xliuhw/NLU-Evaluation-Data}

(3) BANKING77: \url{https://github.com/PolyAI-LDN/task-specific-datasets}

\section{Details for Comparison with State-of-the-arts} 
\label{sec:appendix_main}
\textbf{Comparison Settings.} 
We conduct massive experiments on the three open datasets (TNEWS, HWU64, and BANKING77) for the sake of verifying the effectiveness of our model. ECE is the confidence calibration metric. 

\textbf{Comparison Results.} 
Tabel \ref{tab:main_append} shows more metrics of overall comparison with different calibration methods on the three datasets. 
No matter which dataset we choose, our model achieves the best performance in most metrics among all the calibration methods.
In the case of the TNEWS dataset, though the ACC is a little bit lower than PosCal and LS, the R gains 1.51\% over BERT with the ECE decreases conspicuously.

\section{Details for Performance on Ablation Study}
\label{sec:appendix_ablation}
\textbf{Comparison Settings.}
In order to verify the effectiveness of our components, we gradually removed some components of our model, including hyperspherical space and rebalanced accuracy-uncertainty loss. 
During the practice, we replace the encoded label vector in hyperspherical space with the conventional one-hot label in euclidean space. 

\textbf{Comparison Results.}
Tabel \ref{tab:ablation_append} displays more information on the ablation study on the three open datasets. 
We can clearly see that taking out any components of our model leads to performance reduction, which proves the effectiveness of all components regardless of the dataset we use. 

\section{Details for Performance on TNEWS Dataset with Noise}
\label{sec:appendix_noise}
\textbf{Comparison Settings.}
To verify the effectiveness of our model, we randomly mislabel 5\%, 10\%, 30\%, and 50\% part of the samples' labels on TNEWS dataset to simulate the noisy environment. 

\textbf{Comparison Results.}
The whole information of the performance on TNEWS dataset with noise labels is shown in Table \ref{tab:noise_append}. 
Irrespective of how much noise label is in the dataset, the experimental results indicate that our model still obtains the best performance among all the calibration methods on different metrics. 

\section{Details for Performance of Low-Frequency Labels on TNEWS Dataset}
\label{sec:appendix_LF}
\textbf{Comparison Settings.}
We experiment on the five lowest frequency labels on TNEWS dataset, which possesses fifteen classes in total. 
Table \ref{tab:LF_append} indicates the details of low-frequency labels on TNEWS dataset. 
L-F$_{1}$ means the Lowest Frequency label, L-F$_{2}$ means the second lowest, and so on. 
"Average" means the average performance of the five lowest frequency labels. 

\textbf{Comparison Results.}
As demonstrated in Table \ref{tab:LF_append}, our model achieves the highest average F1 of the lowe-frequency labels, with a 4.06\% absolute increase. 
In addition, it outperforms other calibration methods on the average ECE as well.

\begin{table*}[!t]
  \centering
  \scalebox{0.65}{\setlength{\tabcolsep}{1mm}{
  \begin{tabular}{ccccccccccccccccccccc}
    \toprule[0.9pt]
  \multirow{2.5}{*}{Model} & \multicolumn{5}{c}{5\% Error Labels}                                                & \multicolumn{5}{c}{10\% Error Labels}                                               & \multicolumn{5}{c}{30\% Error Labels}                                               & \multicolumn{5}{c}{50\% Error Labels}                                               \\ \cmidrule(lr){2-6} \cmidrule(lr){7-11} \cmidrule(lr){12-16} \cmidrule(lr){17-21}
                          & ACC            & P              & R              & F1             & ECE            & ACC            & P              & R              & F1             & ECE            & ACC            & P              & R              & F1             & ECE            & ACC            & P              & R              & F1             & ECE            \\  \midrule[0.9pt]
  BERT              & 55.14          & 53.56          & 53.43          & 53.46          & 72.19          & 53.27          & 51.30          & 51.81          & 51.44          & 72.38          & 46.50          & 43.65          & 45.18          & 43.76          & 77.98          & 34.40          & 32.77          & 33.28          & 32.22          & 83.35          \\
  LS                    & 55.04          & 54.10          & 53.91          & 53.95          & 46.46          & 54.26          & 52.62          & 52.63          & 52.59          & 43.81          & 47.19          & 44.34          & 46.35          & 44.83          & 46.92          & 35.19          & 33.35          & 33.94          & 32.91          & 65.94          \\
  PosCal                & 55.14          & 54.64          & 53.76          & 54.11          & 52.77          & 53.78          & 51.90          & 52.49          & 52.14          & 61.92          & 47.98          & 44.96          & 46.74          & 45.40          & 56.83          & 36.18          & 34.25          & 35.43          & 33.93          & 71.37          \\
  AVUC                  & 54.84          & 54.14          & 54.01          & 54.04          & 64.47          & 53.64          & 51.77          & 52.19          & 51.94          & 61.78          & 48.35          & 45.31          & 47.08          & 45.74          & 56.44          & 36.01          & 34.04          & 34.83          & 33.74          & 71.14          \\ \midrule
  HS-RAU                   & \textbf{55.28} & \textbf{54.69} & \textbf{54.30} & \textbf{54.43} & \textbf{34.37} & \textbf{54.52} & \textbf{53.68} & \textbf{53.73} & \textbf{53.66} & \textbf{35.48} & \textbf{48.84} & \textbf{46.53} & \textbf{48.28} & \textbf{47.03} & \textbf{32.25} & \textbf{37.57} & \textbf{35.41} & \textbf{36.80} & \textbf{35.20} & \textbf{41.06}  \\  \bottomrule[0.9pt]
  \end{tabular}}}
  \caption{Performance on TNEWS dataset with noise labels, which are mislabeled randomly.}
  \label{tab:noise_append}
  \end{table*}

\begin{table*}[!t]
  \centering
  \scalebox{0.8}{
  \begin{tabular}{ccccccccccccc}
    \toprule[0.9pt]
  \multirow{2.5}{*}{Model} & \multicolumn{2}{c}{L-F$_{1}$}    & \multicolumn{2}{c}{L-F$_{2}$}    & \multicolumn{2}{c}{L-F$_{3}$}    & \multicolumn{2}{c}{L-F$_{4}$}    & \multicolumn{2}{c}{L-F$_{5}$}    & \multicolumn{2}{c}{Average}     \\  \cmidrule(lr){2-3} \cmidrule(lr){4-5} \cmidrule(lr){6-7} \cmidrule(lr){8-9} \cmidrule(lr){10-11} \cmidrule(lr){12-13}
                         & F1             & ECE            & F1             & ECE            & F1             & ECE            & F1             & ECE            & F1             & ECE            & F1             & ECE            \\  \midrule[0.9pt]
  BERT                   & 40.54          & \textbf{0.338} & 44.50          & 1.817          & 57.54          & 2.686          & 47.25          & 4.277          & \textbf{50.22} & 5.549          & 48.01          & 2.934          \\
  TS                    & 40.54          & 3.185          & 44.50          & 2.864          & 57.54          & 3.096          & 47.25          & 2.804          & \textbf{50.22} & \textbf{3.006} & 48.01          & 2.991          \\
  LS                    & 40.45          & 0.801          & 46.56          & 1.090          & 59.50          & 1.525          & 48.77          & 3.081          & 49.71          & 4.608          & 49.00          & 2.221          \\
  PosCal                & 35.44          & 0.362          & 46.15          & 1.689          & 60.00          & 2.439          & 51.16          & 4.048          & 49.26          & 5.422          & 48.40          & 2.792          \\
  AVUC                  & 43.37          & 0.347          & \textbf{48.31} & 1.640          & \textbf{60.88} & 2.396          & \textbf{51.24} & 4.097          & 49.34          & 5.386          & 50.63          & 2.773          \\  \midrule
  HS-RAU                   & \textbf{54.12} & 1.300          & 47.22          & \textbf{0.965} & 59.06          & \textbf{0.974} & 49.80          & \textbf{2.120} & 50.15          & 3.149          & \textbf{52.07} & \textbf{1.702}  \\ \bottomrule[0.9pt]
  \end{tabular}}
  \caption{Performance of the five lowest frequency labels on TNEWS dataset.}
  \label{tab:LF_append}
  \end{table*}

 \section{More Descriptions}
 Due to space constraints, the theoretical proof is not explained in detail. In fact, previous researches have proved the theory. 
 1) Section 7 of (Szegedy et al. CVPR2016) mentions that soft labels can help mitigate overconfident predictions and improve generalization ability. Minimizing the cross entropy is equivalent to maximizing the log-likelihood of the correct label. So the logit corresponding to the groundtruth label is much great than all other logits. 2) (Hinton et al. NIPS2019) verifies that soft labels do contribute to calibrating the model through experimental analysis. They find that in training, label smoothing drives the activation values into tight clusters, while in the validation set, it propagates around the center and sufficiently covers the predicted confidence range, which reduces ECE. 3) We step further with the soft label method, like LS (label smoothing). Our HS (hyperspherical space) owns a dense label representation matrix, utilizes both the positive and negative regions of space, and distributes more evenly.

\end{document}